\def\year{2021}\relax
\documentclass[letterpaper]{article} 
\usepackage{aaai21}  
\usepackage{times}  
\usepackage{helvet} 
\usepackage{courier}  
\usepackage[hyphens]{url}  
\usepackage{graphicx} 
\usepackage{natbib}  
\usepackage{caption} 
\usepackage{times}
\usepackage{adjustbox}
\usepackage{booktabs}
\usepackage{multirow}
\usepackage{latexsym}
\usepackage{boldline} 

\usepackage{xcolor}
\usepackage{amsfonts}
\usepackage{verbatim}
\usepackage{url}
\usepackage{xspace}
\usepackage{subcaption}
\usepackage{color}
\usepackage{array}
\usepackage{footnote}

\newcommand{\Xt}{{$X$}\xspace}
\newcommand{\Gr}{{$G$}\xspace}
\newcommand{\Cn}{{$C$}\xspace}
\newcommand{\Rp}{{$R$}\xspace}
\newcommand{\CnN}{{$\tilde{C}$}\xspace}
\newcommand{\GrCn}{{$G_C$}\xspace}
\newcommand{\GrCnN}{{$G_{\tilde{C}}$}\xspace}

\newcommand{\GPTia}{{GPT2IA}\xspace}
\newcommand{\CGRG}{{CGRG}\xspace}

\DeclareGraphicsExtensions{.pdf,.png,.jpg}

\usepackage{amssymb}
\usepackage{amsmath}
\usepackage[utf8]{inputenc}
\usepackage{enumitem}
\usepackage[switch]{lineno}

\title{A Controllable Model of Grounded Response Generation}
\author {
        Paper ID: 8469
}
\author{
Zeqiu Wu,\textsuperscript{\rm 1}
Michel Galley,\textsuperscript{\rm 2}
Chris Brockett,\textsuperscript{\rm 2}
Yizhe Zhang,\textsuperscript{\rm 2}
Xiang Gao,\textsuperscript{\rm 2}
Chris Quirk,\textsuperscript{\rm 2}\\
Rik Koncel-Kedziorski,\textsuperscript{\rm 1}
Jianfeng Gao,\textsuperscript{\rm 2}
Hannaneh Hajishirzi,\textsuperscript{\rm 1, 3}
Mari Ostendorf,\textsuperscript{\rm 1}
Bill Dolan \textsuperscript{\rm 2}\\[0.2cm]
}
\affiliations {
    \textsuperscript{\rm 1}University of Washington, Seattle, WA, USA \\
    \textsuperscript{\rm 2}Microsoft Research, Redmond, WA, USA \\
    \textsuperscript{\rm 3}Allen Institute for AI, Seattle, WA, USA \\
    zeqiuwu1@uw.edu, 
    mgalley@microsoft.com
}
\begin{document}
\maketitle
\begin{abstract}
Current end-to-end neural conversation models inherently lack the flexibility to impose semantic control in the response generation process, often resulting in uninteresting responses. Attempts to boost informativeness alone come at the expense of factual accuracy, as attested by pretrained language models' propensity to ``hallucinate'' facts. While this may be mitigated by access to background knowledge, there is scant guarantee of relevance and informativeness in generated responses. We propose a framework that we call controllable grounded response generation (\CGRG), in which lexical control phrases are either provided by a user or automatically extracted by a control phrase predictor from dialogue context and grounding knowledge. Quantitative and qualitative results show that, using this framework, a transformer based model with a novel inductive attention mechanism, trained on a conversation-like Reddit dataset, outperforms strong generation baselines.
\end{abstract}

\section{Introduction}

\begin{figure}
\centering
\includegraphics[width=8.2cm]{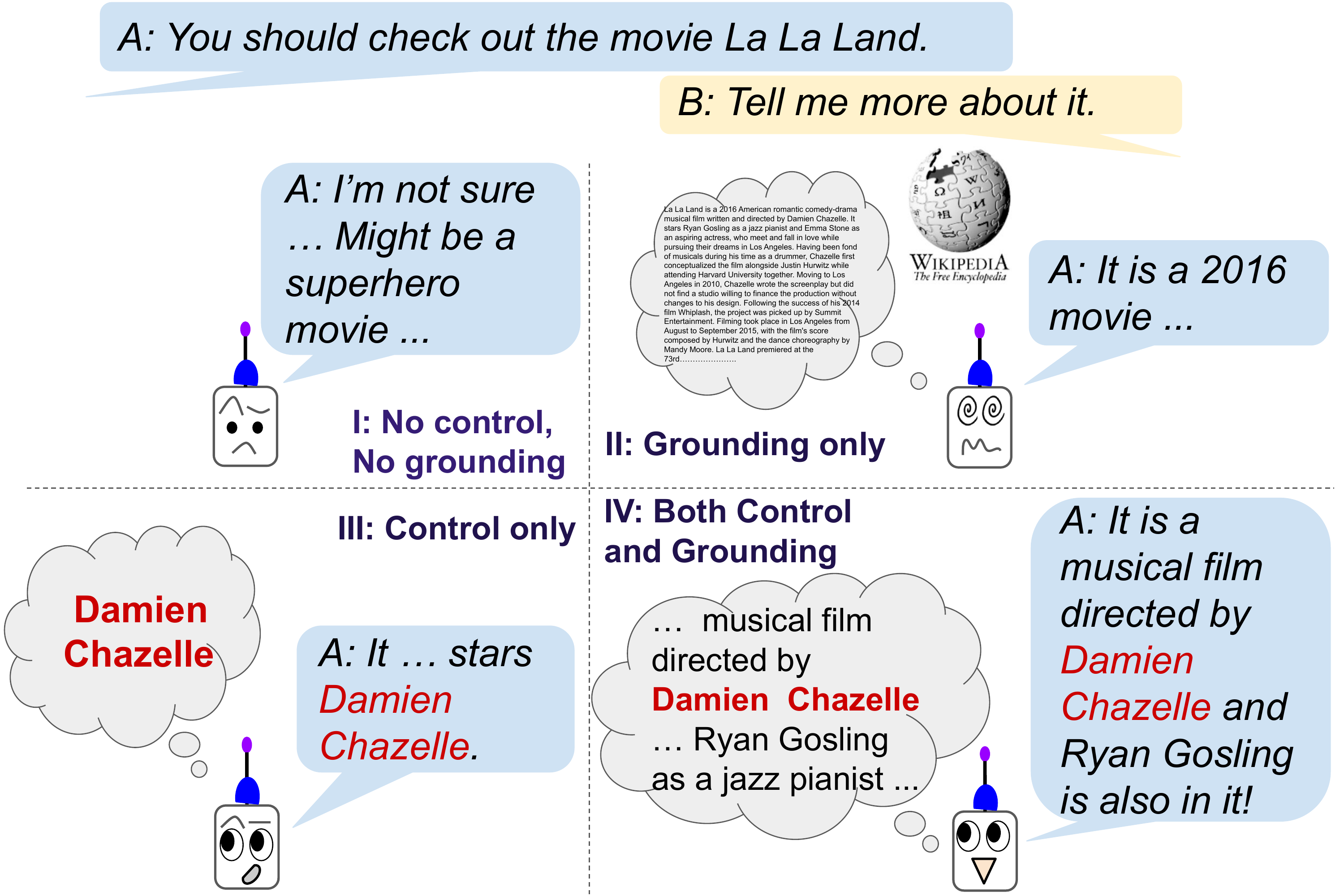}
\caption{\label{fig:teaser1} Generated responses tend to be generic or factually incorrect without grounding or control. Adding grounding improves information reliability but may lead to vague responses. Adding control boosts response specificity, 
but using both leads to contentful and reliable responses.}
\end{figure}

End-to-end neural models for open-domain response generation \citep{shang2015neural,sordoni2015neural,vinyals2015neurcon,gao2019survey} are capable of generating conversational responses that are both fluent and contextually appropriate. Although the earliest neural generation models were characterized  
by bland and evasive responses
\cite{LiEtAl:2016},
surprisingly human-like conversations can be generated using recent diversity-enhancing strategies \cite{holtzman2019curious, gao2019fusion} and 
massive GPT-2 style models \cite{Radford2019gpt2,zhang2019dialogpt}.\footnote{For a related task (document creation), 72\% of human judges found GPT-2 credible vs. 83\% for New York Times articles: \url{https://openai.com/blog/gpt-2-6-month-follow-up/}} 
While blandness may no longer present a challenge, the downside has been a propensity towards ``hallucinated'' or ``fake'' output \cite{zellers2019neuralfakenews} of the kind illustrated in scenario I in  Figure~\ref{fig:teaser1}. %

Grounded response generation \cite{Ghazvininejad2018AKN,Dinan2018WizardOW,Qin2019CMR} approaches can inhibit hallucination of facts. Yet grounding alone (e.g, the Wikipedia page about \textit{La La Land} in scenario II of Figure~\ref{fig:teaser1})  without control and semantic targeting may induce output that is accurate but vague or irrelevant. Controllable text generation \cite{hokamp2017lcd,Keskar2019CTRL,tang2019target, see2019control} provides a level of semantic control that can guide the decoder towards relevant output, but in the absence of grounding the model is prevented from associating control phrases with correct facts. %
We posit that both grounding knowledge and lexical control are essential to generating reliable information. %
We therefore introduce a generation framework called controllable grounded response generation that incorporates both components.  
Lexical controls not only enforce response specificity, but filter lengthy, irrelevant or incoherent groundings.%

\begin{figure}
\centering

\includegraphics[width=8.2cm]{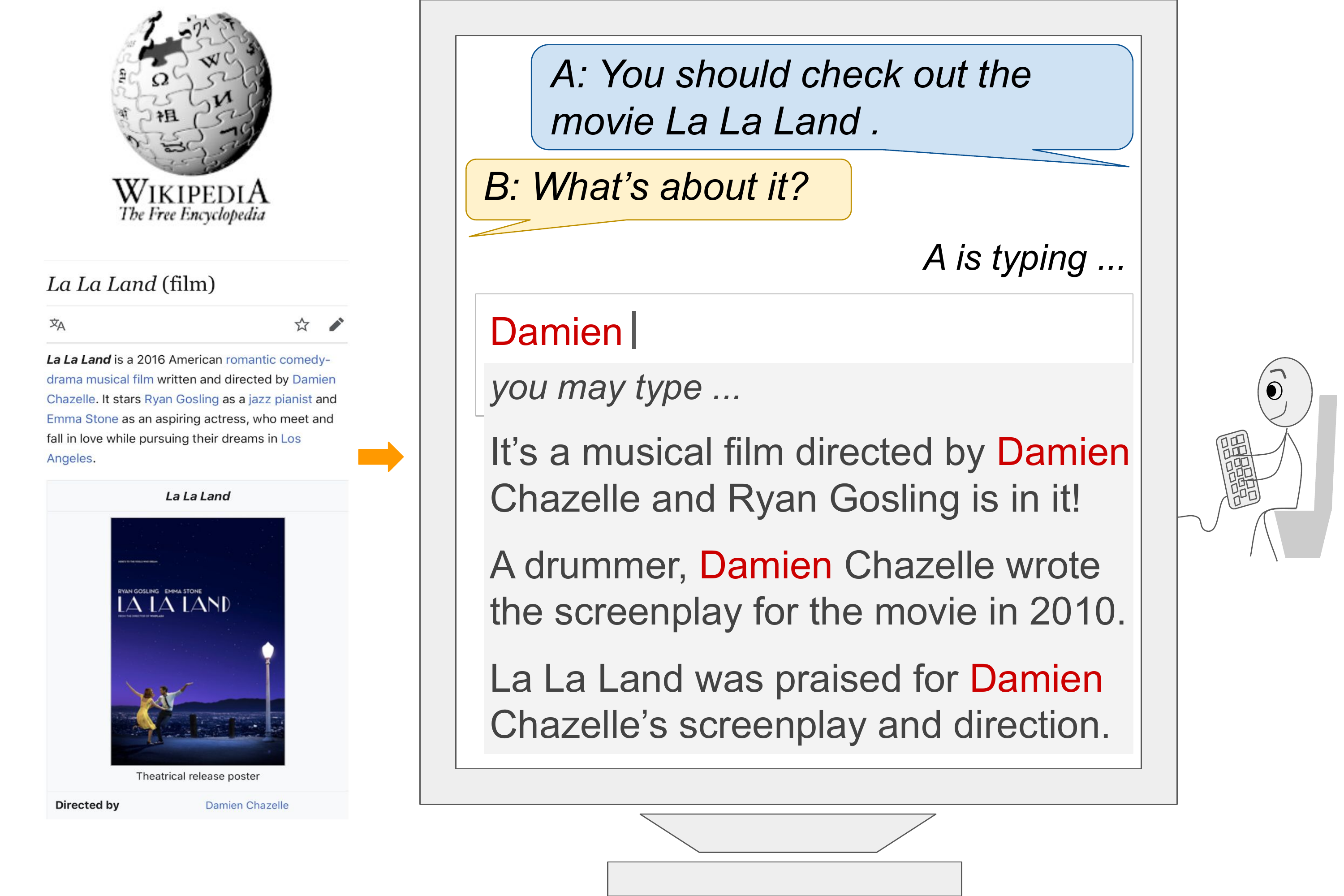}
\caption{\label{fig:teaser2} The machine acts as a response editorial assistant that suggests candidate responses for the user A according to the conversation history, the user's partial input (\textit{Damien}) and grounding knowledge.}
\end{figure}

We consider two scenarios for lexical control of conversational text generation.
Control can come from a human user, as in applications where an editorial assistant helps a person write a 
message.
Figure~\ref{fig:teaser2} depicts a person typing keywords to indicate their semantic intent, while the machine helps construct the response to be sent out.
Alternatively, control could be predicted in a fully automated system.

This work makes the following contributions:
(1)~We propose a novel framework called controllable grounded response generation (\CGRG) that generates a response from the dialogue context, lexical control phrases and groundings. 
To the best of our knowledge, this is the first work to integrate both control and grounding into response generation, and to explore how they can be mutually beneficial. (2)~We combine recent success in transformer-based generation models and a novel inductive attention mechanism to this problem setting. (3)~We show through %
qualitative and quantitative evaluations %
that \CGRG outperforms strong baselines where: a) the control phrases are provided by a (simulated) user, 
and b) automatically extracted by a control phrase prediction model.

\section{Approach}

We formalize the problem as follows: given dialogue context \Xt{}, 
$p$ lexical control phrases \mbox{\Cn{} $= (C_1, \cdots, C_p)$} and
$q$ sentences of grounding \mbox{\Gr{} $= (G_1, \cdots, G_q)$},
generate a response \mbox{\Rp{} $= (r_1, \cdots, r_m)$} that contains semantic information guided by \Cn{}. Control phrases can be either directly provided by a user or automatically derived from a control phrase predictor. 
The \CGRG\ framework
assumes we have a grounded conversational dataset, such as in \cite{Qin2019CMR}. We assume that each data instance consists of a dialogue context, grounding knowledge and a reference response. To analyze this framework, we define a control mechanism that defines one or more control phrases for each instance. 
The controls are lexical phrases that are relevant to both the target response and some part of the grounding knowledge.

\begin{figure*}
\centering
\includegraphics[width=16cm]{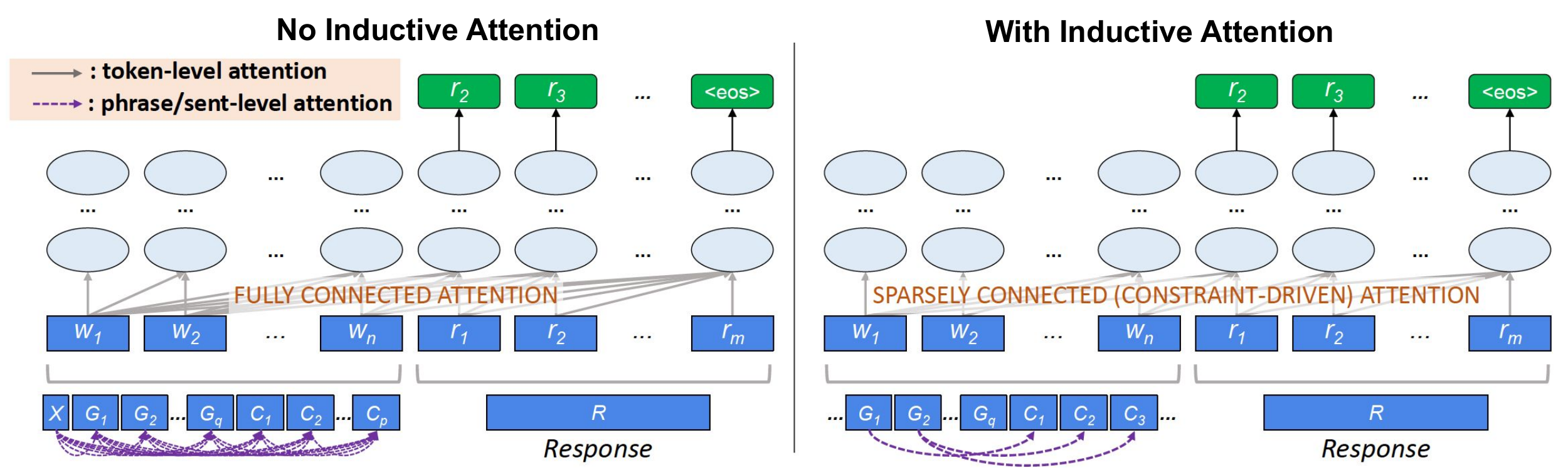}
\caption{\label{fig:gpt2ia} 
Without Inductive Attention, the model considers all possible forward attentions, which can overwhelm the model when the context contains context ($X$), grounding ($G$), and constraints ($C$). In contrast, Inductive Attention uses attentions that are relevant to the constraints. Each dashed arrow applies to all tokens in the corresponding $X$ or $C$ phrase or $G$ grounding.
}
\end{figure*}

To leverage the recent success in transformer-based generation models \footnote{Although our model can be generalized to any attention-based model, we illustrate our method with the basic auto-regressive generation model like GPT-2.} within \CGRG, we concatenate \Xt{}, \Cn 
and \GrCn to be our input sequence, as shown in Figure~\ref{fig:gpt2ia} (left). Then we have the model predict the next response word given the concatenated input sequence (denoted as $S$) and the previous response tokens in \Rp. \GrCn is the subset of \Gr that is relevant to \Cn. For example, in this work, we denote the grounding sentences that contain any phrase in \Cn as \GrCn. 
To differentiate the input elements, we insert an end-of-text token $\langle eos \rangle$ at the end of each dialogue utterance in \Xt{}, a $\langle c \rangle$ token at the end of each control phrase in \Cn and a $\langle s \rangle$ token at the end of each sentence in \GrCn. 

We first concatenate the input sequence $S$ and the response sequence \Rp into a long text. 
We denote the source sequence as $S=(w_1, \cdots , w_n)$, which is used to generate target sentence $R$. The conditional probability of $P(R|S)$ can be written as the product of conditional probabilities:
\begin{align*}\label{eq:lm}
	p(R|S) = \prod_{k=1}^{m+1}  p(r_k|w_1, \cdots, w_{n}, r_1, \cdots, r_{k-1})
\end{align*}
where $r_{m+1}$ is the additional end-of-text token. 

\subsection{Inductive Attention}
\label{sec:gpt2ia}
As shown in Figure~\ref{fig:gpt2ia} (left), a standard transformer-based generation model takes consecutive text sequences as input to train a language model. In our setting, we have input elements \Xt{}, \Cn{}, \GrCn{} in a segmented format. 
Simply concatenating all these input elements can induce noise, as segments may have differential relevance, and we consider attention links between such segments to be uninformative.

We remove potentially uninformative attention links for each data example by injecting pre-established structural information between \Cn and \GrCn. For example, in Figure~\ref{fig:gpt2ia} (right), say that \Cn consists of $C_1, C_2, C_3$, and \GrCn consists of $G_1$ and $G_2$. If we know $C_1$ is only found in 
$G_1$, then we only want to keep the attention link between $C_1$ and $G_1$, and  not between $C_1$ and any of the other grounded sentences. 
Since \GrCn is a set of segmented sentences from \Gr, we remove all cross-sentence links within \GrCn tokens. Similarly, we remove all links between non-identical phrases (e.g., tokens in $C_1$ do not attend to tokens in $C_2$). 
Thus, the attention links for each data example are pre-determined by structural information between \Cn and \GrCn. To implement this, in each transformer layer, we manipulate attention masks by setting undesired attention links to be 0. The others remain at 1. We refer to this pre-calculated attention as inductive attention. Each response token still attends to all input tokens and other response tokens on its left.

We denote the start and end positions of a control phrase $C_i \in C$ in $S$ by $c_i^s$ and $c_i^e$, and those of a grounding sentence $G_j \in G_C$ by $g_j^s$ and $g_j^e$. $G_{C_i}$ denotes the set of grounding sentences in $G_C$ that contain $C_i$. Then we calculate the attention mask $M$ as follows:
\begin{equation*}
M_{a,b} =
\begin{cases}
  0 & \text{if } a < b \\
  0 & \text{if } a \in [c_i^s, c_i^e], b \in [c_{i'}^s, c_{i'}^e], i \neq i'\\
  0 & \text{if } a \in [g_j^s, g_j^e], b \in [g_{j'}^s, g_{j'}^e], j \neq j'\\
  0 & \text{if } a \in [c_i^s, c_i^e], b \in [g_j^s, g_j^e], G_j \not\in G_{C_i}\\
  1 & \text{otherwise }
\end{cases}
\end{equation*}

Then for each transformer head, we have the stacked matrices Q, K and V to represent each example sequence (concatenated $S$ and $T$) as in \cite{vaswani:17}. We calculate the attention as follows ($d$ is the model dimension):  
\begin{align*}\label{eq:attention}
	\text{Attention}(Q, K, V) = \text{softmax} (\frac{M \circ QK^T}{\sqrt{d}})V 
\end{align*}

\subsection{Control Phrase Selection} 
\label{sec:control_phrase}

\paragraph{User Control.} Our user control is defined by lexical phrase(s). Since it is costly to have humans annotate the control phrases associated with an existing set of human comment-response pairs, we use lexical matching to simulate the human-controlled scenario.  Specifically, we define control phrases \Cn{} to be informative n-grams ($n\le 5$) that appear in both the grounding document and the reference response, where informativeness is defined based on a document frequency threshold. 
When two n-grams are identical except for an added function word or punctuation, we use only the shorter version. In addition, we remove the matched n-grams that appear in dialogue context,
with the goal of focusing on providing new information. 
We provide human verification details for such control phrases in Section~\ref{sec:dataset}.

\paragraph{Automatic Control Phrase Predicting.} For the fully automatic scenario, we experiment with two control phrase predictors. We denote these as \CnN{} to differentiate from the user-provided \Cn{}.
The first predictor uses a simple retrieval-based strategy. Given a dialogue context, we rank sentences in \Gr by IDF-weighted word overlaps with \Xt and select the two most frequent n-grams in the top 50 sentences as \CnN. In order to reduce the search space, we use noun phrases only. 

The second predictor leverages the BERT QA system, which is fine-tuned using the dialogue context \Xt as the query, \Gr as the document and the control phrases in \Cn as answers. Then we use the fine-tuned model to predict answers on test examples. We obtain the top two answers as predicted control phrases {\CnN}. 
For both predictors, we drop the second phrase if the string fully overlaps with the first.  

\section{Controllable Conversation Dataset}
\label{sec:dataset}

As a social media aggregator, Reddit is effectively a dataset of multiple domains. We start with the grounded Reddit conversation dataset described in \citet{Qin2019CMR}. This dataset features Reddit conversations about web pages such as news stories and Wikipedia articles, and covers 178 subreddit topics ranging from news/technology to literature/music and multiple writing styles.

We want to focus on contexts where controllable generation is useful. Thus, we keep only responses where at least one matched phrase can be found in the grounding document. Strict lexical matching between target response and grounding assures that the retained examples have a high ratio of grounding utilization. The number of utterances of train, dev and test 
are 390K, 6.7K and 21K, respectively. The average length of all reference responses is 26.5, which is about 40\% longer than in the full dataset due to the focus on controllability. The average numbers of phrases in \Cn for train, dev and test set are 1.32, 1.27 and 1.38 respectively. The average numbers of sentences in \GrCn for train, dev and test set are 4.37, 4.32 and 4.25 respectively. 

To verify that the simulated user control phrases are appropriate for the responses, we use crowd-sourced workers to annotate whether the extracted control phrases are central to the reference response, given the dialogue context. For each response, we had 3 judges to enter a score on a scale of 1 (completely unrelated) to 6 (very central), where 5 was ``somewhat central'' and 4 was ``neutral.'' In 2000 annotated examples, the median score was 4.33 and 67.4\% of examples had a score over 4. Inter-rater agreement was ``fair'' with Krippendorff's alpha coefficient at 0.32.\footnote{This dataset is a filtered version of \cite{Qin2019CMR}'s public dataset. To further facilitate reproducibility, we release our data preparation and modeling code at \url{https://github.com/ellenmellon/CGRG}.}

\section{Experimental Setup}

\subsection{Training and Inference Setup}

In our experiments, all transformer-based models have both type and positional embedding for each input token. We treat \Xt, each sentence in \GrCn, each phrase in \Cn and response \Rp as separate segments. We set the maximum number of sentences in \GrCn to be 20 and maximum number of phrases in \Cn to be 10, then we have ``0'' for \Xt; ``1-20'' for \GrCn; ``21-30'' for \Cn and ``31'' for \Rp tokens as type embedding. For each segment, we have the position embedding for each token as its position in that segment. 

We use GPT-2 as the basic transformer-based generation model in our experiments for drawing fair comparison with the DialoGPT \citep{zhang2019dialogpt} architecture, a state-of-the-art conversational response generation model trained on 147M Reddit comment chains on the basis of GPT-2. We use the small version of GPT-2 with 117M parameters, with the maximum length of the input or target response sequence to be 512. 
We use BPE tokenization, following GPT-2.
We finetune our model and all other GPT-2-based baselines (including DialoGPT) on our controllable and grounded Reddit dataset on top of the original DialoGPT.  None of their Reddit training or validation examples overlap with our test examples. We use batch size 32. Learning rate (1e-5) and warmup steps (1600) are tuned on 
the dev set perplexity, with all other parameters being the same as DialoGPT \footnote{All model configurations in Section~\ref{evaluated-systems} share the same hyperparameter values (only tuned 3 times on top of the original values from DialoGPT) except CMR, which has the same parameters in the original paper.}. Each training process is run on 2 Tesla K-80 nodes.

We use greedy search as the decoding strategy for all GPT-2 and GPT2IA setups, except for a single experiment setting where grid beam search (GBS) \citep{hokamp2017lcd} is applied for comparison with lexically constrained decoding. %
We also compare our methods with GBS to investigate whether it helps to encode the constraints into the hidden state during both training and inference, as GBS uses lexical constraints only during inference.

\subsection{Evaluated Systems}
\label{evaluated-systems}

\paragraph{Models}
Although our designed model can be applied to any transformer-based generation model, we use GPT-2 as our base model for experiments. We use the following models for experiments: (a) $\textbf{GPT-2}$, which has the same architecture as DialoGPT \citep{zhang2019dialogpt}, under the input setting $\textbf{\Xt}$ (see below) or the first line in both Table~\ref{gold-cstr} and Table~\ref{noisy-cstr}; (b) $\textbf{GPT2IA}$ model with inductive attention; (c) $\textbf{GPT-2 + GBS}$ that applies the attended GPT-2 model, while control phrases $C$ are given to the model at decoding time;\footnote{Such constrained decoding is based on grid beam search (GBS) introduced in \citet{hokamp2017lcd}, where lexical control phrases are added in decoding only, without involving training.} and (d) $\textbf{CMR}$  \citep{Qin2019CMR}, which is a previous state-of-the-art grounded response generation model on the Reddit dataset that combines a MRC model and an LSTM decoder.

\paragraph{Input Settings} We evaluate the above models according to the following settings to analyze how control and grounding help improve the response generation performance:

\begin{itemize}[topsep=5pt, partopsep=0pt, itemsep=-1pt,leftmargin=9pt]
\item
$\textbf{\Xt:}$ This is the standard setting for non-controllable response generation, where only the dialogue context is given. 
We conduct experiments for the GPT-2 generation model. Note that GPT-2 in this setting is the same as the DialoGPT architecture.
\item
$\textbf{\Xt{}+\Gr:}$ This is the standard setting for grounded response generation. We compare two models: CMR and GPT-2. GPT-2 for this setting concatenates \Xt and \Gr as its input. As both models have an input sequence length limit, only a random subset of grounding is fed into each model.
\item
$\textbf{\Xt{}+\Cn:}$ This is the controllable response generation setting without grounding.
We conduct GPT-2 experiments by concatenating \Xt and \Cn.
\item
$\textbf{\Xt{}+\GrCn:}$ This setting measures how the grounding only relevant to \Cn can help with response generation, without explicitly providing \Cn. We conduct experiments for GPT-2, by concatenating \Xt and \GrCn as the input.
\item
$\textbf{\Xt{}+\Cn{}+\GrCn:}$ This setting measures how grounded control can help with response generation. We conduct experiments for GPT-2 and GPT2IA, by concatenating \Xt, \GrCn and \Cn as the input.
\item
$\textbf{\Xt{}+\Cn{}+\Gr:}$ This setting is for comparison against existing constrained generation methods like grid beam search (GBS) introduced in \citet{hokamp2017lcd}, where lexical control phrases are added in decoding only without involving training. We conduct experiments for GPT-2 where \Xt and \Gr are the only encoded inputs and \Cn is only applied in decoding with GBS.
\end{itemize}

\smallskip
\subsection{Automatic Evaluation}
\label{sec:eval}

Previous work \citep{li2016persona, sun2019ssim} has shown that automatic metrics for generation can be unreliable and have low absolute values, so we rely on human evaluation for our main conclusions.  However, due to the high cost of human evaluation, automatic evaluation metrics can be useful for hyper-parameter tuning and model selection.
For automatic evaluation, we measure the relevance of the generated responses with three metrics: BLEU-4 \citep{Papineni2002bleu}, 
NIST-4 \citep{Doddington2002nist}  
(a variant of BLEU that weights n-gram matches by their information gain, penalizing uninformative n-grams), and diversity of bigrams in generated responses (Div-2, the ratio between the number of distinct vs.\ total bigrams). 
The human evaluation is used to verify the improvements of the best case systems as determined by the automatic metrics.

We experiment with both user-controllable and fully automatic response generation, with simulated user-selected and predicted lexical control phrases, respectively. As different reference responses correspond to different ``gold'' control phrases, we use single-reference evaluation for the user-controllable setting. 
Predicted control phrases are independent of reference responses, so we use multi-reference evaluation.
Comments in Reddit discussions are often associated with multiple responses, which provide a multi-reference test set.
For each metric, we report the highest score among up to 5 alternative human references.

\subsection{Human Evaluation}
\label{sec:human_eval}

Human evaluation was conducted using crowd-sourced workers. 
Judges were presented with paired randomized outputs.
The document title, a short snippet of the document and up to two conversational turns were provided as context.
Relevance and appropriateness to the preceding dialog and consistency with the background text (as a metric of factual correctness) were measured.
Judgments were based on a five-point Likert scale, and ties were permitted. Three to four judges evaluated each pair, and metrics were imposed to block poorly performing judges. Inter-rater agreement was ``fair'' with Krippendorff's coefficient at 0.32.\footnote{Sample sizes vary. The number was reduced from an initial 1,000 when we automatically removed a number of instances where egregiously offensive content rendered them inappropriate to display to judges.}  

\section{Results and Analysis}

\subsection{User-controlled Response Generation}
\label{sec:gold-cstr}

The user-controllable grounded response generation experiments are summarized in Table~\ref{gold-cstr}, using single-reference evaluation. 
The low BLEU/NIST scores are consistent with differences between human references as seen in Sec.~\ref{sec:planner-results}.

\begin{savenotes}
\begin{table}
\footnotesize
\centering
\begin{tabular}{@{\hskip3pt}l@{\hskip3pt}|@{\hskip3pt}l@{\hskip3pt}|@{\hskip3pt}r@{\hskip5pt} r@{\hskip5pt}|@{\hskip3pt}r@{\hskip3pt}|@{\hskip3pt}c@{\hskip3pt}@{\hskip3pt}}
\hline \textbf{Setting} & \textbf{Model} & \textbf{NIST} & \textbf{BLEU} & \textbf{Div-2} & \textbf{Avg-L}\\ \hline
1) \Xt & GPT-2 & 0.90 & 0.55\% & 4.9\% & 22.2 \\
2) \Xt{}+\Gr & CMR & 0.34 & 0.17\% & 11.3\% & 15.1 \\	
3) \Xt{}+\Gr & GPT-2 & 0.98 & 0.67\% & 7.5\% & 23.1 \\\hline \hline

4) \Xt{}+\Cn & GPT-2 & 1.67 & 2.65\% & 10.7\% & 28.7 \\
5) \Xt{}+\GrCn & GPT-2 & 1.34 & 1.58\% & 11.1\% & 26.6 \\
6) \Xt{}+\Cn{}+\Gr & GPT-2+GBS\footnote{\Xt{}+\Cn{}+\Gr (GBS) only takes \Xt{}+\Gr as the encoder input while \Cn is seen at decoding only.} & 1.60 & 2.38\% & 10.6\% &  26.8 \\ \hline
7) \Xt{}+\Cn{}+\GrCn & GPT-2 & 1.77 & 3.22\% & 11.3\% & 27.0 \\  
8) \Xt{}+\Cn{}+\GrCn & GPT2IA & \textbf{1.80} & \textbf{3.26}\% & \textbf{11.6}\% & 25.9 \\
\hline
\end{tabular}
\caption{\label{gold-cstr}User-controllable Response Generation automatic evaluation.}
\end{table}
\end{savenotes}
\begin{table}
\normalsize
\centering

\begin{tabular}{@{}rr|r|rl@{}}
\cmidrule[\heavyrulewidth]{1-5} 
\multicolumn{2}{r|}{GPT2IA} & \multicolumn{1}{c|} {Tied} & \multicolumn{2}{l}{GPT-2}\\ 
\cmidrule[\heavyrulewidth]{1-5}
\multicolumn{5}{c}{\textbf{Relevance}: \textit{Which response is more relevant}}\\
\multicolumn{5}{c}{\textit{and appropriate to the preceding dialog?}}\\
\cmidrule[\heavyrulewidth]{1-5} 
\Xt{}+\Cn{}+\GrCn & \textbf{69.8}\% & 14.1\% &  16.1\% & \Xt{}+\Cn{}+\Gr{}+GBS\\
\Xt{}+\Cn{}+\GrCn & \textbf{42.1}\% & 23.5\% & 34.4\% &  \Xt{}+\Cn{}\\
\Xt{}+\Cn{}+\GrCn & \textbf{38.1}\% & 28.6\%  & 33.3\% & \Xt{}+\Cn{}+\GrCn\\
\cmidrule[\heavyrulewidth]{1-5}
\multicolumn{5}{c}{\textbf{Consistency}: \textit{Which response is more}}\\
\multicolumn{5}{c}{\textit{consistent with the grounding text?} }\\
\cmidrule[\heavyrulewidth]{1-5} 
\cmidrule{1-5}
\Xt{}+\Cn{}+\GrCn & 28.1\% & 44.3\% & 27.6\% &  \Xt{}+\Cn{}+\GrCn\\
\Xt{}+\Cn{}+\GrCn & \textbf{37.6}\% & 31.4\% & 31.0\% & \Xt{}+\Cn{}\\
\cmidrule[\heavyrulewidth]{1-5}
\end{tabular}

\caption{Controllable Response Generation human evaluation for 
relevance and background consistency, showing preferences (\%). A number in bold indicates that the system is significantly better at $p \leq 10^{-5}$, computed using 10k bootstrap replications. 
}
\label{tab:human_eval}
\end{table}

Lines 1-3 are not controllable settings, while lines 4-8 have control phrases as input, either explicitly or implicitly. The performance gap between lines (1-3) and (4-8) demonstrates the value of adding control.
Additionally, we can draw the following conclusions by comparing rows in Table~\ref{gold-cstr}: (i)~\mbox{\textbf{1 vs. 3}}: Simply adding grounding to the model input improves the performance somewhat;
(ii) \textbf{\mbox{2 vs. 3}}: GPT-2 in general performs better than the LSTM-based model CMR, indicating that the combination of pre-training and having a transformer-based decoder helps improve generation; 
(iii) \textbf{\mbox{3 vs. 5}}: providing constraint-sensitive
grounding boosts 
performance compared to having all the grounding
(iv) \textbf{\mbox{5 vs. 7-8}}: providing control phrases in an explicit way is important; (v)~\textbf{\mbox{6 vs. 7-8}}: applying control in hidden states helps the model generate better quality responses than applying control at decoding only; (vi) \textbf{\mbox{7 vs. 8}}: inductive attention helps reduce noise and improve the performance.  

Comparing lines \textbf{\mbox{6 vs. 7-8}} we see that applying control in hidden states is more effective than strict constraints at decoding, but it is possible that controls at the training and decoding stages could be complementary. Investigation of methods of combining these are left to future research.

Human evaluation results in Table~\ref{tab:human_eval} confirm that \Xt{}+\Cn{}+\GrCn{}+\GPTia outperforms other systems, except in the case of Consistency, where there is no statistical difference between \Xt{}+\Cn{}+\GrCn{}+\GPTia and \Xt{}+\Cn{}+\GrCn{}+GPT2, both grounded systems.

\subsection{Predictor-controlled Response Generation} 
\label{sec:planner-results}

In the fully automatic response generation scenario, we compare two models for predicting control phrases. 
Table~\ref{noisy-cstr} compares the two models to the setting where no control phrases are provided to the model, using multi-reference evaluation. We observe that both the retrieval-based and BERT QA based control phrase predictors outperform \Xt{}+GPT-2 (DialoGPT) and achieve good Div-2 results, with NIST scores similar to the user-controlled settings but low BLEU score. 
For more insight into automatic scores, we also report scores on human responses. When defining the multi-reference responses, we hold out one response in each set as the ``human'' system setting. These human responses have higher diversity, but their NIST and BLEU scores remain low owing to the huge range of possible responses to any comment.

In paired comparisons using human judges our fully automatic system \Xt{}+\CnN{}+\GrCnN+\GPTia was rated as having the most informative and relevant response in 32\% of cases, significantly better than the 20-21\% for \Xt{}+GPT-2 (DialoGPT).

\begin{table} %
\footnotesize
\centering
\begin{tabular}{@{\hskip3pt}l@{\hskip3pt}|@{\hskip3pt}l@{\hskip3pt}|@{\hskip3pt}l@{\hskip3pt}|@{\hskip3pt}r@{\hskip5pt}r@{\hskip3pt}|@{\hskip3pt}r@{\hskip3pt}}
\hline \textbf{Setting} & \textbf{Model} & \textbf{Phrase Predictor} & \textbf{NIST} & \textbf{BLEU} & \textbf{Div-2} \\ \hline

\Xt & GPT-2 & - & 1.42  & \textbf{1.31}\% & 18.1\% \\ 
\Xt{}+\GrCnN & GPT-2 & Retrieval-based & 1.61 & 1.26\% & 19.4\% \\
\Xt{}+\CnN{}+\GrCnN & GPT2IA & Retrieval-based & \textbf{1.67} & 1.23\%  & \textbf{20.2}\% \\
\Xt{}+\CnN{}+\GrCnN & GPT2IA & BertQA & \textbf{1.67} & 1.26\% & 19.6\% \\ \hline
Human & - & - & 2.04 & 2.56\% & 62.8\% \\
\hline
\end{tabular}
\caption{\label{noisy-cstr}Response Generation automatic evaluation (multi-references) using constraints from control phrase predictor. Note that results of Tables~\ref{gold-cstr}~and~\ref{noisy-cstr}, as user constraints give away significant information about the intended response.}
\end{table}

\subsection{Qualitative Analysis} 

\begin{figure*}
\centering
\fontsize{9pt}{9pt}\selectfont
\begin{subfigure}[t]{1\linewidth}
\centering
{
\begin{tabular}{@{}p{1\textwidth}@{}}\\
{\bf Grounding ($G_C$):} Sam got his bachelor degree in Physics at University of Science and Technology of China. He spent 6 months at University of Tokyo in Japan as a visiting student, when he was a master student in Computer Science at University of Hong Kong from 2010-2012. And he finished his PhD at University of Toronto in Canada with his research focused on interpretability of neural networks on text generation in 2017.\\
{\bf Context ($X$):} Do you know the education background of the new faculty, Sam?\\
{\bf Control phrases ($C$):} University of Toronto; Neural networks\\
{\bf Model predictions:}
\end{tabular}
}
\end{subfigure}\vspace{0.1cm}
\begin{subfigure}[t]{0.49\linewidth}
\centering
\includegraphics[width=1.0\linewidth]{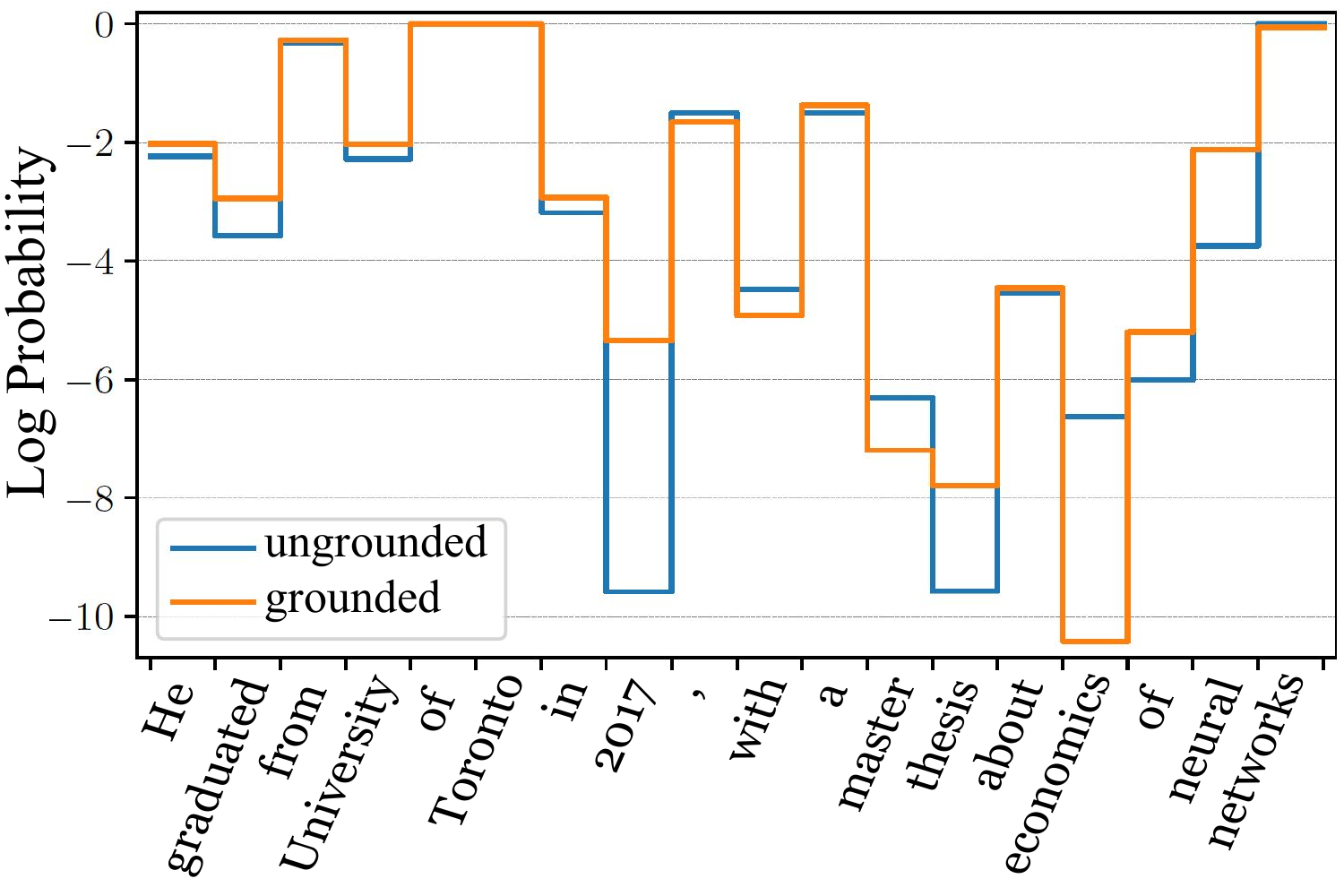}
\caption{\label{fig:token_prob} The grounded model (\Xt{}+\Cn{}+\GrCn{}+GPT2IA) offers better discrimination vis-\`a-vis an ungrounded model (\Xt{}+\Cn{}+GPT2), given a document about a person's education background.
} 
\end{subfigure}
\hfill
\begin{subfigure}[t]{0.49\linewidth}
\centering
\includegraphics[width=1.0\linewidth]{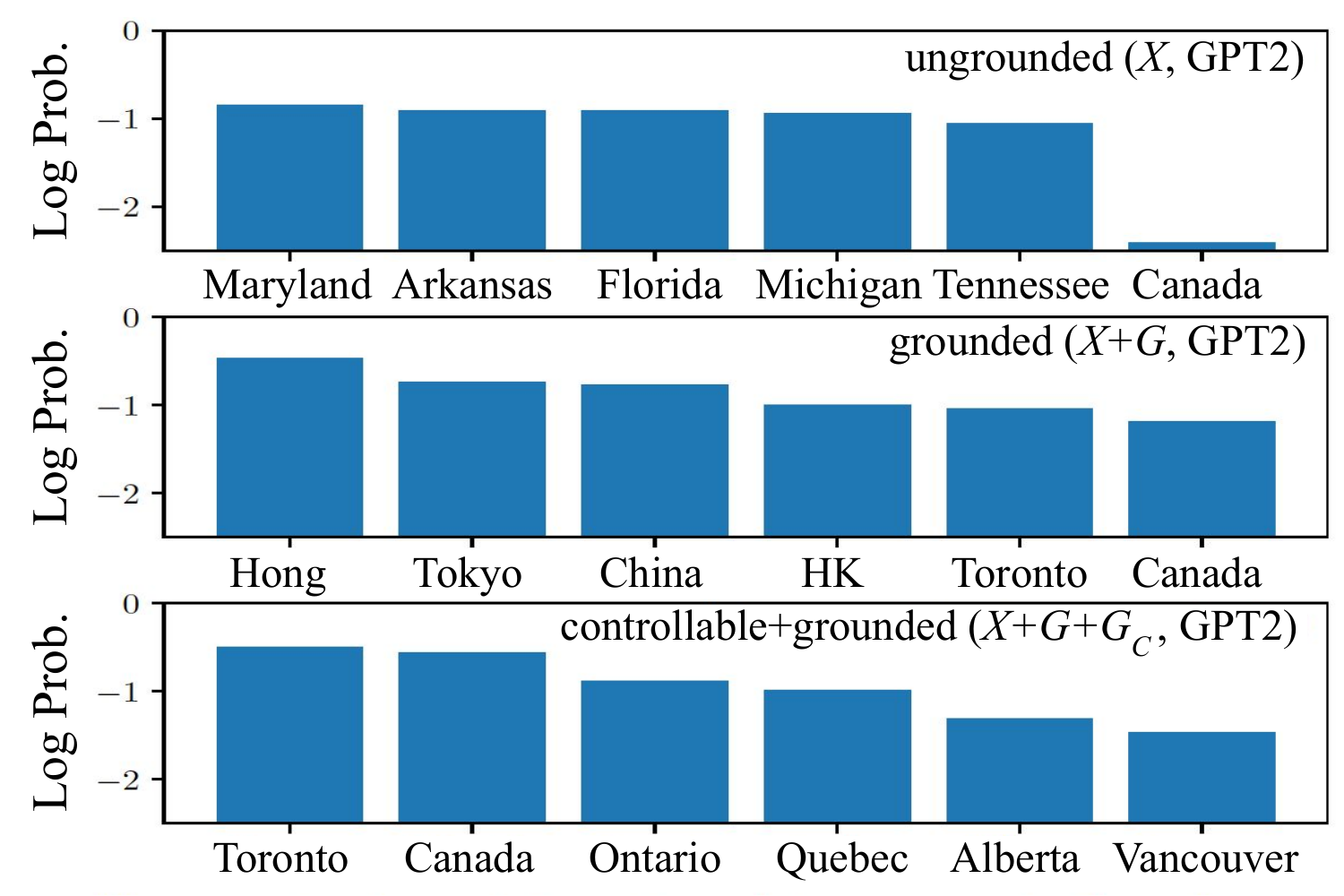}
\caption{\label{fig:top_tokens} The top 5 tokens (plus \textit{Canada}) generated after the partial response \textit{Sam just graduated from University of}. The ungrounded model prefers generic predictions. The grounded model is more topically relevant. The constraint further positively influences the hidden state.}
\end{subfigure}
\caption{Effect of grounding and control on text generation.}
\end{figure*}



To understand how grounding knowledge assists generation, we plot the token-level probability (Figure~\ref{fig:token_prob}) for both \Xt{}+\Cn{} and \Xt{}+\Cn{}+\GrCn systems. We intentionally select an example about an uncommon entity to eliminate the possibility that the knowledge is captured in pre-training. 
The figure shows the token-level probability of a potential response, 
given a dialogue context, two control phrases, and grounding sentences.
The grounded model assigns higher probabilities to contextual words from grounding such as \textit{graduated} and \textit{thesis} and to factually correct entity tokens like \textit{2017}. It assigns lower probability to factually incorrect tokens like \textit{economics}. These observations suggest that grounding knowledge can help controllable generation: contextualize control phrases and  distinguish correct vs.\ incorrect facts.

Figure~\ref{fig:top_tokens} illustrates the functions of control and grounding. We list the top 6 tokens after a partial response given the same dialogue context and grounding, and control phrase \textit{Canada}. The ungrounded non-controllable model gives equally distributed probabilities to well-known American state names after \textit{University of}. Adding grounding helps the model rank locations based on background knowledge. Further adding controls helps the model locate the correct or intended answer.%

To quantify the observations in Figure~\ref{fig:token_prob} and Figure~\ref{fig:top_tokens}, we sample 100 test examples and randomly pick an entity from each reference response to calculate the entity's probability from each model. We restrict the entity to be non-occurring in control phrases. Then we calculate the average probability ratio for \Xt{}+\Cn{}/\Xt{}+\Cn{}+\GrCn and \Xt{}+\Gr/\Xt{}+\Cn{}+\GrCn, to be 0.773 and 0.886 respectively. Both of them are smaller than 1.0, which indicates having both grounding and control phrases gives higher probability to correct entities than either of these alone. %
Explicit control phrases can be leveraged to dissect the generation process. 
Table~\ref{case-study} shows how controls may guide or perturb the GPT2IA model to produce responses with diverging semantics. 
An example with sample outputs of different systems in the user-controlled scenario is shown in Table~\ref{case-study-all-examples}.

\begin{table*}
\centering
\normalsize
\begin{tabular}{p{1.6cm}|p{12cm}}

\hline 
Dialogue Context & With ``nihonium'', Japanese scientists become first from an Asian country to name atomic element.\\
\hline\hline
Control & \textbf{periodic table} \\
Grounding & ... The \textbf{periodic table} is a great legacy in chemistry ...\\
\Xt{}+\Cn{}+\GrCn +GPT2IA & I'm not sure if this is a good thing or not, but I'm pretty sure the \textbf{periodic table} is a great legacy in chemistry.\\
\hline
Control & \textbf{artificially} \\
Grounding & ... The \textbf{artificially} synthesized element has 113 protons in its nucleus ...\\
\Xt{}+\Cn{}+\GrCn +GPT2IA & I wonder if they will be able to name a chemical that \textbf{artificially} produces atomic elements.\\

\hline
\end{tabular}
\caption{\label{case-study} For the same dialogue context, GPT2IA generates varied responses given different control phrases.}
\end{table*}

\begin{table*}
\centering
\normalsize
\begin{tabular}{p{1.6cm}|p{12cm}}
\hline 
Context & 76 \% of all known serial killers in the 20th century were from the United States.\\
Control &  \textbf{law enforcement}\\
Grounding & ... and may include more stringent policies for military personnel in \textbf{law enforcement} or security ... Should the cases cross multiple jurisdictions, the \textbf{law enforcement} system in \textbf{the US} is fragmented and thus not configured to detect multiple similar murders across a large geographic area ...\\
\hline
\Xt &  I'm pretty sure \textbf{the US} had a police force of around 100,000 people.\\
\Xt{}\Cn & I'm pretty sure the USA had a large number of serial killers in \textbf{the US}. I'm sure the USA had a large number of \textbf{law enforcement} officers in \textbf{the US}.\\
\Xt{}\Cn{}\GrCn &  I'm not sure if this is true, but I'm pretty sure that \textbf{the US} has a lot of \textbf{law enforcement} officers that are from \textbf{the US}.\\
\Xt{}\Cn{}\GrCn+IA &  I'm not sure if this is true, but I'm pretty sure that the \textbf{law enforcement} in \textbf{the US} is not very good at \textbf{detecting} serial killers.\\

\hline
\end{tabular}
\caption{\label{case-study-all-examples} Sample outputs of the systems, with baseline outputs for comparison.}
\vspace{-2mm}
\end{table*}

\section{Related Work}

\subsection{Grounded Response Generation} 
Although some relevant work draws on external knowledge sources, none incorporates user control. \citet{Ghazvininejad2018AKN} develop a memory network based model that leverages grounding information from Foursquare tips. \citet{Moghe2018TowardsEB} and \citet{zhou2018dataset} collect movie discussion datasets via crowdsourcing. These are limited to specific domains. \citet{Dinan2018WizardOW} crowdsource conversations where each utterance is grounded in up to one single sentence. We focus on a more realistic, scalable setting in which a response may constitute a blend of multiple grounding 
informations, 
rather than a single factual sentence rephrasing. Other researchers propose a copy mechanism to import tokens from 
dialogue context and grounding \citep{Yavuz2018DeepCopy} or leverage a reading comprehension model to co-encode dialogue context and grounding \citep{Qin2019CMR}. 

Other work incorporates relational knowledge bases \citep{zhu2017flexible, liu2018diffusion} or commonsense knowledge graphs \citep{young2018augment} to conversational models. More recently, \citet{liu2019kgconv} develop a graph-path-based method on knowledge graphs augmented with unstructured grounding. Our present work focuses on text based grounding and does not require preconstructed knowledge graphs.

\subsection{Controlled and Content-Planned Generation} 

Prior work on machine translation and language generation 
has sought to enforce user-specified constraints, primarily in the form of lexical constraints \citep{hokamp2017lcd,hu2019ParaBank,hu-etal-2019-improved,miao2019cgmh}. These approaches exploit constraints at inference time only; in our case, constraints are applied during training, with the option also of application at inference. Application during training enables the constraints to be incorporated into the latent space for better predictions.

In other related work, \cite{see2019control,Keskar2019CTRL,tang2019target} have explored non-lexical constraints, but do not examine how these could facilitate use of grounding and external knowledge. We see this line of research as complementary to ours. These papers also make the assumption that (gold) constraints can always given to the system, which limits the potential to demonstrate broader benefits of the approaches. To address this concern, we also evaluate our models in settings where {\it gold} constraints are unavailable (e.g., based on predicted constraints produced by a control phrase predictor).

Controllable text generation has also been employed in text style transfer \cite{hu2017toward} and other tasks \cite{ficler2017controlling,dong2017learning,gao2019structuring}, to disentangle high-level style information from contextual information such that the former can be independently manipulated. \cite{zhao2018unsupervised} uses discrete latent actions to learn an interpretable representation for task-oriented dialogue systems.
While these works use ``style'' labels (e.g. positive/negative, formal/informal) as controlling signals, our framework controls generation with specific lexical constraints, allowing for fine-grained semantic control.%

Content planned generation \citep{Hua2019content, Wiseman2017content} targets selection of a few keyphrases or table entries as the focus of text generation. However this line of work does not need to consider dialogue context, which is essential for response generation.
\section{Conclusion}

The \CGRG framework allows users to inject soft semantic control into the generation process. It incorporates grounding to contextualize users' semantic intents as well as to boost information reliability. We introduce an inductive attention mechanism for self-attention-based generation models like GPT-2 in order to boost its performance. We also demonstrate that this framework can benefit standard automatic response generation when integrated with a control phrase predictor.
Some interesting future directions include exploring various types of user desired control and extending the controllable grounded generation concept to broader generation tasks like document writing assistance.

\section*{Acknowledgements}
We thank members of Microsoft Research and University of Washington's NLP groups who provided feedback and insights to this work.

\bibliography{aaai21}

\begin{thebibliography}{39}
\providecommand{\natexlab}[1]{#1}
\providecommand{\url}[1]{\texttt{#1}}
\providecommand{\urlprefix}{URL }
\expandafter\ifx\csname urlstyle\endcsname\relax
  \providecommand{\doi}[1]{doi:\discretionary{}{}{}#1}\else
  \providecommand{\doi}{doi:\discretionary{}{}{}\begingroup
  \urlstyle{rm}\Url}\fi

\bibitem[{Dinan et~al.(2019)Dinan, Roller, Shuster, Fan, Auli, and
  Weston}]{Dinan2018WizardOW}
Dinan, E.; Roller, S.; Shuster, K.; Fan, A.; Auli, M.; and Weston, J. 2019.
\newblock Wizard of {W}ikipedia: Knowledge-Powered Conversational agents.
\newblock In \emph{ICLR}.

\bibitem[{Doddington(2002)}]{Doddington2002nist}
Doddington, G. 2002.
\newblock Automatic evaluation of machine translation quality using n-gram
  cooccurrence statistics.
\newblock In \emph{Proc. of HLT}, 138--145.

\bibitem[{Dong et~al.(2017)Dong, Huang, Wei, Lapata, Zhou, and
  Xu}]{dong2017learning}
Dong, L.; Huang, S.; Wei, F.; Lapata, M.; Zhou, M.; and Xu, K. 2017.
\newblock Learning to generate product reviews from attributes.
\newblock In \emph{Proc. of EACL}, 623--632.

\bibitem[{Ficler and Goldberg(2017)}]{ficler2017controlling}
Ficler, J.; and Goldberg, Y. 2017.
\newblock Controlling Linguistic Style Aspects in Neural Language Generation.
\newblock In \emph{Proc. of EMNLP}, 94--104.

\bibitem[{Gao, Galley, and Li(2019)}]{gao2019survey}
Gao, J.; Galley, M.; and Li, L. 2019.
\newblock Neural approaches to conversational {AI}.
\newblock \emph{Foundations and Trends in Information Retrieval} 13(2-3):
  127–298.

\bibitem[{Gao et~al.(2019{\natexlab{a}})Gao, Lee, Zhang, Brockett, Galley, Gao,
  and Dolan}]{gao2019fusion}
Gao, X.; Lee, S.; Zhang, Y.; Brockett, C.; Galley, M.; Gao, J.; and Dolan, B.
  2019{\natexlab{a}}.
\newblock Jointly Optimizing Diversity and Relevance in Neural Response
  Generation.
\newblock In \emph{Proc. of NAACL}, 1229--1238.

\bibitem[{Gao et~al.(2019{\natexlab{b}})Gao, Zhang, Lee, Galley, Brockett, Gao,
  and Dolan}]{gao2019structuring}
Gao, X.; Zhang, Y.; Lee, S.; Galley, M.; Brockett, C.; Gao, J.; and Dolan, B.
  2019{\natexlab{b}}.
\newblock Structuring Latent Spaces for Stylized Response Generation.
\newblock In \emph{Proc. of EMNLP}, 1814--1823.

\bibitem[{Ghazvininejad et~al.(2018)Ghazvininejad, Brockett, Chang, Dolan, Gao,
  tau Yih, and Galley}]{Ghazvininejad2018AKN}
Ghazvininejad, M.; Brockett, C.; Chang, M.-W.; Dolan, B.; Gao, J.; tau Yih, W.;
  and Galley, M. 2018.
\newblock A Knowledge-Grounded Neural Conversation Model.
\newblock In \emph{Proc. of AAAI}, 5110--5117.

\bibitem[{Hokamp and Liu(2017)}]{hokamp2017lcd}
Hokamp, C.; and Liu, Q. 2017.
\newblock Lexically Constrained Decoding for Sequence Generation Using Grid
  Beam Search.
\newblock In \emph{Proc. of ACL}, 1535--1546.

\bibitem[{Holtzman et~al.(2020)Holtzman, Buys, Forbes, and
  Choi}]{holtzman2019curious}
Holtzman, A.; Buys, J.; Forbes, M.; and Choi, Y. 2020.
\newblock The Curious Case of Neural Text Degeneration.
\newblock In \emph{ICLR}.

\bibitem[{Hu et~al.(2019{\natexlab{a}})Hu, Khayrallah, Culkin, Xia, Chen, Post,
  and Van~Durme}]{hu-etal-2019-improved}
Hu, J.~E.; Khayrallah, H.; Culkin, R.; Xia, P.; Chen, T.; Post, M.; and
  Van~Durme, B. 2019{\natexlab{a}}.
\newblock Improved Lexically Constrained Decoding for Translation and
  Monolingual Rewriting.
\newblock In \emph{Proc. of NAACL}, 839--850.

\bibitem[{Hu et~al.(2019{\natexlab{b}})Hu, Rudinger, Post, and
  Durme}]{hu2019ParaBank}
Hu, J.~E.; Rudinger, R.; Post, M.; and Durme, B.~V. 2019{\natexlab{b}}.
\newblock {ParaBank}: Monolingual Bitext Generation and Sentential Paraphrasing
  via Lexically-constrained Neural Machine Translation.
\newblock In \emph{Proc. of AAAI}, 6521--6528.

\bibitem[{Hu et~al.(2017)Hu, Yang, Liang, Salakhutdinov, and
  Xing}]{hu2017toward}
Hu, Z.; Yang, Z.; Liang, X.; Salakhutdinov, R.; and Xing, E.~P. 2017.
\newblock Toward Controlled Generation of Text.
\newblock In \emph{Proc. of ICML}, 1587--1596.

\bibitem[{Hua and Wang(2019)}]{Hua2019content}
Hua, X.; and Wang, L. 2019.
\newblock Sentence-Level Content Planning and Style Specification for Neural
  Text Generation.
\newblock In \emph{Proc. of EMNLP}, 591--602.

\bibitem[{Keskar et~al.(2019)Keskar, McCann, Varshney, Xiong, and
  Socher}]{Keskar2019CTRL}
Keskar, N.~S.; McCann, B.; Varshney, L.~R.; Xiong, C.; and Socher, R. 2019.
\newblock {CTRL}: A Conditional Transformer Language Model for Controllable
  Generation.
\newblock \emph{Computing Research Repository} arXiv:1909.05858.
\newblock Version 2.

\bibitem[{Li et~al.(2016{\natexlab{a}})Li, Galley, Brockett, Gao, and
  Dolan}]{LiEtAl:2016}
Li, J.; Galley, M.; Brockett, C.; Gao, J.; and Dolan, B. 2016{\natexlab{a}}.
\newblock A Diversity-Promoting Objective Function for Neural Conversation
  Models.
\newblock In \emph{Proc. of NAACL}, 110--119.

\bibitem[{Li et~al.(2016{\natexlab{b}})Li, Galley, Brockett, Spithourakis, Gao,
  and Dolan}]{li2016persona}
Li, J.; Galley, M.; Brockett, C.; Spithourakis, G.; Gao, J.; and Dolan, B.
  2016{\natexlab{b}}.
\newblock A persona-based neural conversation model.
\newblock In \emph{Proc. of ACL}, 994--1003.

\bibitem[{Liu et~al.(2018)Liu, Chen, Ren, Feng, Liu, and
  Yin}]{liu2018diffusion}
Liu, S.; Chen, H.; Ren, Z.; Feng, Y.; Liu, Q.; and Yin, D. 2018.
\newblock Knowledge Diffusion for Neural Dialogue Generation.
\newblock In \emph{Proc. of ACL}, 1489--1498.

\bibitem[{Liu et~al.(2019)Liu, Niu, Wu, and Wang}]{liu2019kgconv}
Liu, Z.; Niu, Z.-Y.; Wu, H.; and Wang, H. 2019.
\newblock Knowledge Aware Conversation Generation with Explainable Reasoning
  over Augmented Graphs.
\newblock In \emph{Proc. of EMNLP}, 1782--1792.

\bibitem[{Miao et~al.(2019)Miao, Zhou, Mou, Yan, and Li}]{miao2019cgmh}
Miao, N.; Zhou, H.; Mou, L.; Yan, R.; and Li, L. 2019.
\newblock {CGMH:} Constrained Sentence Generation by Metropolis-Hastings
  Sampling.
\newblock In \emph{Proc. of AAAI}, 6834--6842.

\bibitem[{Moghe et~al.(2018)Moghe, Arora, Banerjee, and
  Khapra}]{Moghe2018TowardsEB}
Moghe, N.; Arora, S.; Banerjee, S.; and Khapra, M.~M. 2018.
\newblock Towards Exploiting Background Knowledge for Building Conversation
  Systems.
\newblock In \emph{Proc. of EMNLP}, 2322--2332.

\bibitem[{Papineni et~al.(2002)Papineni, Roukos, Ward, and
  Zhu}]{Papineni2002bleu}
Papineni, K.; Roukos, S.; Ward, T.; and Zhu, W. 2002.
\newblock {BLEU}: A Method for Automatic Evaluation of Machine Translation.
\newblock In \emph{Proc. of ACL}, 311--318.

\bibitem[{Qin et~al.(2019)Qin, Galley, Brockett, Liu, Gao, Dolan, Choi, and
  Gao}]{Qin2019CMR}
Qin, L.; Galley, M.; Brockett, C.; Liu, X.; Gao, X.; Dolan, B.; Choi, Y.; and
  Gao, J. 2019.
\newblock Conversing by Reading: Contentful Neural Conversation with On-demand
  Machine Reading.
\newblock In \emph{Proc. of ACL}, 5427--5436.

\bibitem[{Radford et~al.(2019)Radford, Wu, Child, Luan, Amodei, and
  Sutskever}]{Radford2019gpt2}
Radford, A.; Wu, J.; Child, R.; Luan, D.; Amodei, D.; and Sutskever, I. 2019.
\newblock Language Models are Unsupervised Multitask Learners.
\newblock \emph{OpenAI Blog.} Accessed 22 March 2021.

\bibitem[{See et~al.(2019)See, Roller, Kiela, and Weston}]{see2019control}
See, A.; Roller, S.; Kiela, D.; and Weston, J. 2019.
\newblock What makes a good conversation? How controllable attributes affect
  human judgments.
\newblock In \emph{Proc. of NAACL}, 1702--1723.

\bibitem[{Shang, Lu, and Li(2015)}]{shang2015neural}
Shang, L.; Lu, Z.; and Li, H. 2015.
\newblock Neural Responding Machine for Short-Text Conversation.
\newblock In \emph{Proc. of ACL-IJCNLP}, 1577--1586.

\bibitem[{Sordoni et~al.(2015)Sordoni, Galley, Auli, Brockett, Ji, Mitchell,
  Nie, Gao, and Dolan}]{sordoni2015neural}
Sordoni, A.; Galley, M.; Auli, M.; Brockett, C.; Ji, Y.; Mitchell, M.; Nie,
  J.-Y.; Gao, J.; and Dolan, B. 2015.
\newblock A neural network approach to context-sensitive generation of
  conversational responses.
\newblock In \emph{Proc. of NAACL-HLT}, 196--205.

\bibitem[{Sun and Nenkova(2019)}]{sun2019ssim}
Sun, S.; and Nenkova, A. 2019.
\newblock The Feasibility of Embedding Based Automatic Evaluation for Single
  Document Summarization.
\newblock In \emph{Proc. of EMNLP}, 1216--1221.

\bibitem[{Tang et~al.(2019)Tang, Zhao, Xiong, Liang, Xing, and
  Hu}]{tang2019target}
Tang, J.; Zhao, T.; Xiong, C.; Liang, X.; Xing, E.~P.; and Hu, Z. 2019.
\newblock Target-Guided Open-Domain Conversation.
\newblock In \emph{Proc. of ACL}, 5624--5634.

\bibitem[{Vaswani et~al.(2017)Vaswani, Shazeer, Parmar, Uszkoreit, Jones,
  Gomez, Kaiser, and Polosukhin}]{vaswani:17}
Vaswani, A.; Shazeer, N.; Parmar, N.; Uszkoreit, J.; Jones, L.; Gomez, A.~N.;
  Kaiser, L.; and Polosukhin, I. 2017.
\newblock Attention is All you Need.
\newblock In \emph{Advances in Neural Information Processing Systems 30},
  5998--6008.

\bibitem[{Vinyals and Le(2015)}]{vinyals2015neurcon}
Vinyals, O.; and Le, Q. 2015.
\newblock A Neural Conversational Model.
\newblock In \emph{Proc. of ICML Deep Learning Workshop}.

\bibitem[{Wiseman, Shieber, and Rush(2017)}]{Wiseman2017content}
Wiseman, S.; Shieber, S.; and Rush, A. 2017.
\newblock Challenges in data-to-document generation.
\newblock In \emph{Proc. of EMNLP}, 2253--2263.

\bibitem[{Yavuz et~al.(2019)Yavuz, Rastogi, Chao, and
  Hakkani-Tur}]{Yavuz2018DeepCopy}
Yavuz, S.; Rastogi, A.; Chao, G.-L.; and Hakkani-Tur, D. 2019.
\newblock {DeepCopy}: Grounded Response Generation with Hierarchical Pointer
  Networks.
\newblock In \emph{Proc. of SIGdial Meeting on Discourse and Dialogue},
  122--132.

\bibitem[{Young et~al.(2018)Young, Cambria, Chaturvedi, Huang, Zhou, and
  Biswas}]{young2018augment}
Young, T.; Cambria, E.; Chaturvedi, I.; Huang, M.; Zhou, H.; and Biswas, S.
  2018.
\newblock Augmenting end-to-end dialogue systems with commonsense knowledge.
\newblock In \emph{Proc. of AAAI}, 4970--4977.

\bibitem[{Zellers et~al.(2019)Zellers, Holtzman, Rashkin, Bisk, Farhadi,
  Roesner, and Choi}]{zellers2019neuralfakenews}
Zellers, R.; Holtzman, A.; Rashkin, H.; Bisk, Y.; Farhadi, A.; Roesner, F.; and
  Choi, Y. 2019.
\newblock Defending Against Neural Fake News.
\newblock In \emph{NeurIPS}, 9051--9062.

\bibitem[{Zhang et~al.(2020)Zhang, Sun, Galley, Chen, Brockett, Gao, Gao, Liu,
  and Dolan}]{zhang2019dialogpt}
Zhang, Y.; Sun, S.; Galley, M.; Chen, Y.-C.; Brockett, C.; Gao, X.; Gao, J.;
  Liu, J.; and Dolan, B. 2020.
\newblock {DialoGPT}: Large-Scale Generative Pre-training for Conversational
  Response Generation.
\newblock In \emph{Proc. of ACL System Demonstrations}, 270–278.

\bibitem[{Zhao, Lee, and Eskenazi(2018)}]{zhao2018unsupervised}
Zhao, T.; Lee, K.; and Eskenazi, M. 2018.
\newblock Unsupervised Discrete Sentence Representation Learning for
  Interpretable Neural Dialog Generation.
\newblock In \emph{Proc. of ACL}, 1098--1107.

\bibitem[{Zhou, Prabhumoye, and Black(2018)}]{zhou2018dataset}
Zhou, K.; Prabhumoye, S.; and Black, A.~W. 2018.
\newblock A Dataset for Document Grounded Conversations.
\newblock In \emph{Proc. of EMNLP}, 708--713.

\bibitem[{Zhu et~al.(2017)Zhu, Mo, Zhang, Zhu, Peng, and
  Yang}]{zhu2017flexible}
Zhu, W.; Mo, K.; Zhang, Y.; Zhu, Z.; Peng, X.; and Yang, Q. 2017.
\newblock Flexible End-to-End Dialogue System for Knowledge Grounded
  Conversation.
\newblock \emph{Computing Research Repository} arXiv:1709.04264.

\end{thebibliography}

\end{document}